\documentclass[conference]{IEEEtran}
\IEEEoverridecommandlockouts
\usepackage{cite}
\usepackage{amsmath,amssymb,amsfonts}
\usepackage{algorithmic}
\usepackage{graphicx}
\usepackage{textcomp}
\usepackage{xcolor}
\usepackage[numbers]{natbib}
\def\BibTeX{{\rm B\kern-.05em{\sc i\kern-.025em b}\kern-.08em
    T\kern-.1667em\lower.7ex\hbox{E}\kern-.125emX}}

\usepackage{array}
\newcolumntype{P}[1]{>{\centering\arraybackslash}p{#1}}   
\usepackage{hyperref} 

\begin{document}

\title{Resolving Camera Position for a Practical Application of Gaze Estimation on Edge Devices
\thanks{This work was partly supported by the National Research Foundation of Korea (NRF) grant funded by the Korea government (MSIT) (No. 2019R1A2C2087489), and Ministry of Culture, Sports and Tourism(MCST) and Korea CreativeContent Agency(KOCCA) in the Culture Technology(CT) Research \& Development (R2020070004) Program 2021.}
} 

\author{\IEEEauthorblockN{Linh Van Ma, Tin Trung Tran, Moongu Jeon}
\IEEEauthorblockA{\textit{School of Electrical Engineering and Computer Science} \\
\textit{Gwangju Institute of Science and Technology}\\
Gwangju, South Korea \\
\{linh.mavan, ttrungtin, mgjeon\}@gist.ac.kr}
}

\maketitle

\begin{abstract}
Most Gaze estimation research only works on a setup condition that a camera perfectly captures eyes gaze. They have not literarily specified how to set up a camera correctly for a given position of a person. In this paper, we carry out a study on gaze estimation with a logical camera setup position. We further bring our research in a practical application by using inexpensive edge devices with a realistic scenario. That is, we first set up a shopping environment where we want to grasp customers gazing behaviors. This setup needs an optimal camera position in order to maintain estimation accuracy from existing gaze estimation research. We then apply the state-of-the-art of few-shot learning gaze estimation to reduce training sampling in the inference phase.  In the experiment, we perform our implemented research on NVIDIA Jetson TX2 and achieve a reasonable speed, 12 FPS which is faster compared with our reference work, without much degradation of gaze estimation accuracy. The source code is released at \href{https://github.com/linh-gist/GazeEstimationTX2}{\color{blue}{https://github.com/linh\/-gist/GazeEstimationTX2}}. 
\end{abstract}


\section{Introduction}
Businesses are now benefiting from computer vision applications. One of the well-known businesses in the USA, Amazon Go \cite{wankhede2018just} has successfully applied deep learning, sensor fusion, and computer vision for checkout, purchase, and proceed for payment automatically without any human interactions. Hence, understanding customers' attention/behavior/preferences during shopping is crucial to increase avenues. For example, a customer who buys milk usually looks for bread. We can place milk and bread next to each other.  Gaze is an individual's perception and awareness of human visual attention. We can track the gaze to know which products customers prefer the most and their shopping behaviors.

Many eye commercial trackers perfectly measure the motion of an eye relative to the head. For example, Tobii \cite{gibaldi2017evaluation} can estimate and track eye gaze without requiring recalibration. However, it is expensive and hard to upgrade once a business decides to buy those physical devices. More importantly, upgradeable, inexpensive, and easy-to-deploy are the key points to leverage business avenues.

In this paper, we introduce research on nearly real-time eye gaze estimation. We use low-priced RGB cameras to capture eyes gaze frames.  Those frames are processed by economical edge devices to find out where a person stares at. More specifically, we first reflect a shopping environment where a customer looks at a shelf to buy goods, groceries. We then find an optimal position inside the shelf to set up an RGB camera. This position supports the camera appropriately grasping the customer's eyes gaze. Afterward, we build a deep eye gaze estimation network for edge devices. In this network, images captured from an RBG camera are first fed into a face detection module. Subsequently, facial landmark detection is employed to find key points on facial images. Finally, gaze estimation is accurately estimated from those obtained key points. We use few-shot learning to reduce the number of samples require to fine-tune a deep gaze estimation network. This few-shot learning also makes our work more easily to be deployed in a real-world application because we can fine-tune our deep network within a few samples. In the experiment, we perform our implemented system on NVIDIA Jetson TX2. We achieve a reasonable speed, 12 FPS which is faster compared with our reference work, without much degradation of gaze estimation accuracy. We also prove that the camera should be set up in an optimal position to increase gaze estimation accuracy.

This paper is organized as follows. Section \ref{related} presents our based research. In Section \ref{system}, we logically find an optimal camera position that supports the camera to accurately estimate a person's eyes gaze. Subsequently, we demonstrate our method with several experiments in Section \ref{experiment}. Finally, we conclude this paper with a future research direction. 

\section{Related Works}
\label{related}

In gaze estimation, we first need to detect faces using face detectors \cite{viola2004robust,zhang2016joint, farfade2015multi,schroff2015facenet}. MTCNN proposed in \cite{zhang2016joint} is a fast and efficient facial detection model. MTCNN composes of deep cascaded networks, Proposal Network (P-Net), Refinement Network (R-Net), and Output Network (O-Net). These three networks exploit the properties correlation between face alignment (in R-Net) and face detection (in P-Net) to increase the performance of face detection (in O-Net). Furthermore, the authors propose a new online hard sample mining strategy leading to an improvement during the training process with fewer manual factors.

Facial landmarks \cite{kazemi2014one,sun2019high} to find key points (68 landmarks) on detected faces, is the next crucial step for gaze estimation. Facial landmark detectors essentially try to label and localize the seven facial regions as follows: (1) Right eyebrow, (2) Left eyebrow, (3) Right eye, (4) Left eye, (5) Nose, (6) Mouth, and (7) Jaw. More specifically, the authors \cite{kazemi2014one} propose a general framework based on gradient boosting for learning an ensemble of regression trees that optimizes the sum of square error loss and naturally handles missing or partially labeled data. In short, an ensemble regression trees is trained to estimate the face's landmark positions directly from a sparse subset of pixel intensities. Notably, their result can achieve face alignment in milliseconds for a single image. This result brings us a chance to implement facial landmark detection on edge devices where we do not have many computational resources. Fortunately, this module was well implemented in Dlib \cite{king2009dlib}. In contrast, in \cite{sun2019high} the authors propose a HRNetV2, a modification on HRNet \cite{sun2019deep}, to work high-resolution representation learning. This learning leads to stronger representations and higher landmark localization accuracy. Hence, we use this HRNetV2 to detect facial landmarks while fine-tuning our gaze estimation network. It has high accuracy but not fast compared to \cite{kazemi2014one}. 


Gaze estimation \cite{fischer2018rt, park2018deep} is a process to predict where a person is gazing at given an image with the person's full face. Similar to our approach \cite{park2019few} argued that each person has a distinct gaze, a person-specific gaze. It leads to limit the accuracy of person-independent gaze estimation networks. Hence, they propose personalizing gaze networks. This network encodes face appearance, gaze direction, and head rotation into latent space by using a disentangling encoder-decoder architecture. Their method allows the network to learn person-specific gaze within a few samples (less than nine samples).

Thanks to a real-time detecting eye blink algorithm proposed in \cite{cech2016real}, we can determine whether eyes are widely opened or slightly closed. They first use landmark positions to calculate the eye aspect ratio (EAR). A Support Vector Machine \cite{noble2006support} classifier is subsequently employed to determine whether eyes are blinking or non-blinking pattern based on EAR value. In our approach, this EAR allows us to check whether our camera position set up on the shelf is optimal or not because eyes are slightly narrower when a person's eyes look downward. Given a camera position, we can check this EAR value to determine the position can avoid the above effect or not.

In \cite{sagonas2013300,sagonas2013semi}, the authors argue that the limitation of accuracy from facial landmark detection comes from the training process, which is lack of quality and quantity of annotated face databases. The databases mostly were manually annotated by a trained expert and the fatigue factor is hard to avoid which lead to the error during the works. Hence, Sagonas et al. \cite{sagonas2013300,sagonas2013semi} propose a unified annotation pipeline with a semi-automatic annotation system. They use Active Orientation Models (AOMs) generative models \cite{tzimiropoulos2014active} to train their network with an image from mixed expression and viewing angle. The resulted train model can generate accurate annotation in different conditions and can be generalized to unseen images. Along with the problem of the database, the authors \cite{zhang2018revisiting} further introduce a data normalization method to combine the images and gaze direction to a normalized space, which can cancel out the varieties of head and eye pose positions.


\section{System Overview}
\label{system}

\begin{figure}\begin{center}
\includegraphics[width=0.487\textwidth]{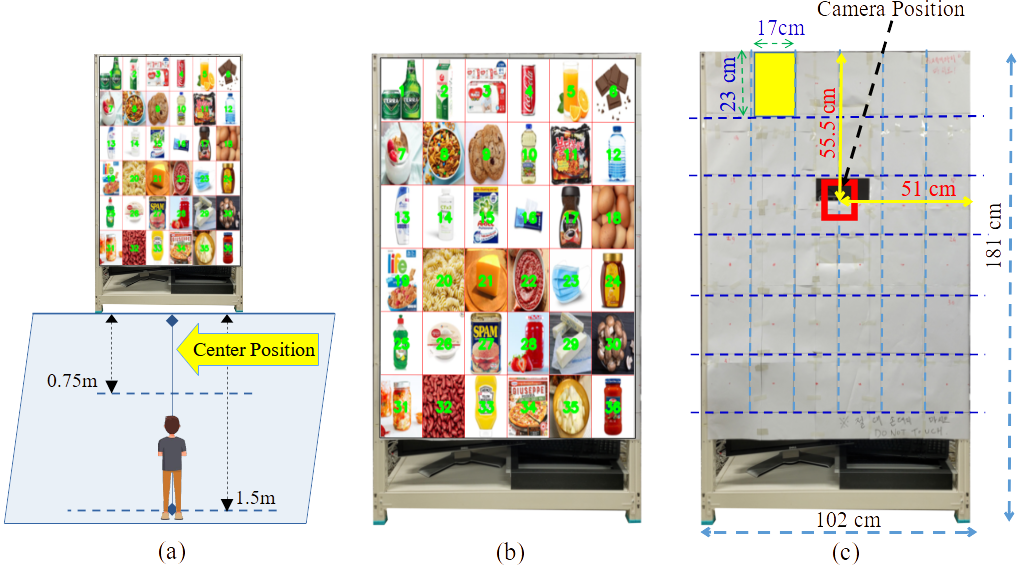}
\caption{(a) We ideally model a shelf in a store, (b) each item is mapped to a labeled rectangle number from the top-left to the bottom-right, (c) a simplified version of a shelf in a store.} \label{fig1}\end{center}
\end{figure}

We ideally model a physical store in our research environment with a shelf as shown in Fig.~\ref{fig1}. It has goods and groceries placed separately in different positions on the shelf. In Fig.~\ref{fig1} (a), a user stands closely 0.75 meters and far at most 1.5 meters from the shelf. In Fig.~\ref{fig1} (b), we divide the shelf into 36 (6x6) rectangles labeled from 1 at the top-left to 36 at the bottom right. Each rectangle has a size of 17 centimeters in width and 23 centimeters in height. Each item is mapped/encoded to one or two rectangles. To simplify, in Fig.~\ref{fig1} (c), we remove groceries and put a large paper with printed 36 rectangles onto the shelf. The camera is located somewhere inside the shelf behind the large white paper. If a unit measurement is not specified, we use centimeters throughout our paper.

\begin{figure}\begin{center}
\includegraphics[width=0.487\textwidth]{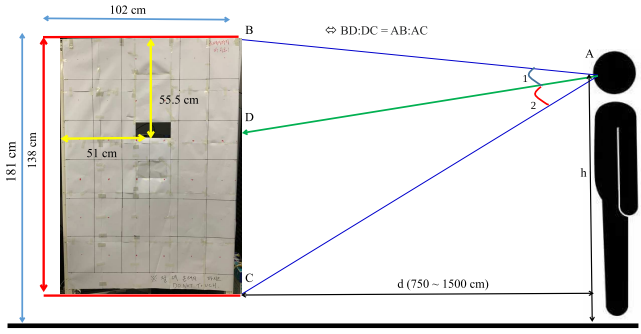}
\caption{A shelf with 181 (cm) height and width of 102 (cm). A person stands away from the shelf with a distance ranging from 750 to 1500 (cm). The camera places optimally at the center width and 55.5 (cm) from the top. (A) represents the person's eyes (or middle of them), (B) top of the shelf, (C) bottom of the large white paper, not floor, (D) is the optimal camera position on the self.} \label{fig2}\end{center}
\end{figure}

Fig.~\ref{fig2} illustrates a side view projection of Fig.~\ref{fig1} (a) (projection from left to right). We name $B$ is at the top, $C$ is at the bottom (of the large white paper, not the floor), D is at the camera location on the shelf. $A$ locates at the person's eyes (or middle of them). From the eyes to $B$, $D$, and $C$, we have two angles $\alpha_1 = \widehat{BAD}$, $\alpha_2 = \widehat{DAC}$. The camera should be in the optimal position so that when a person's head rotates vertically (up and down, or pitch in Euler angle), $AD$ should equally divide the angle $\widehat{BAC}$ created when the person looks at the top and bottom of the shelf as shown in Fig.~\ref{fig2}. In other words, the camera should make this equality $\alpha_1=\alpha_2$ happens. A person looks at the shelf while standing in the range 750 to 1500 centimeters far from the shelf. 

\begin{figure}\begin{center}
\includegraphics[width=0.487\textwidth]{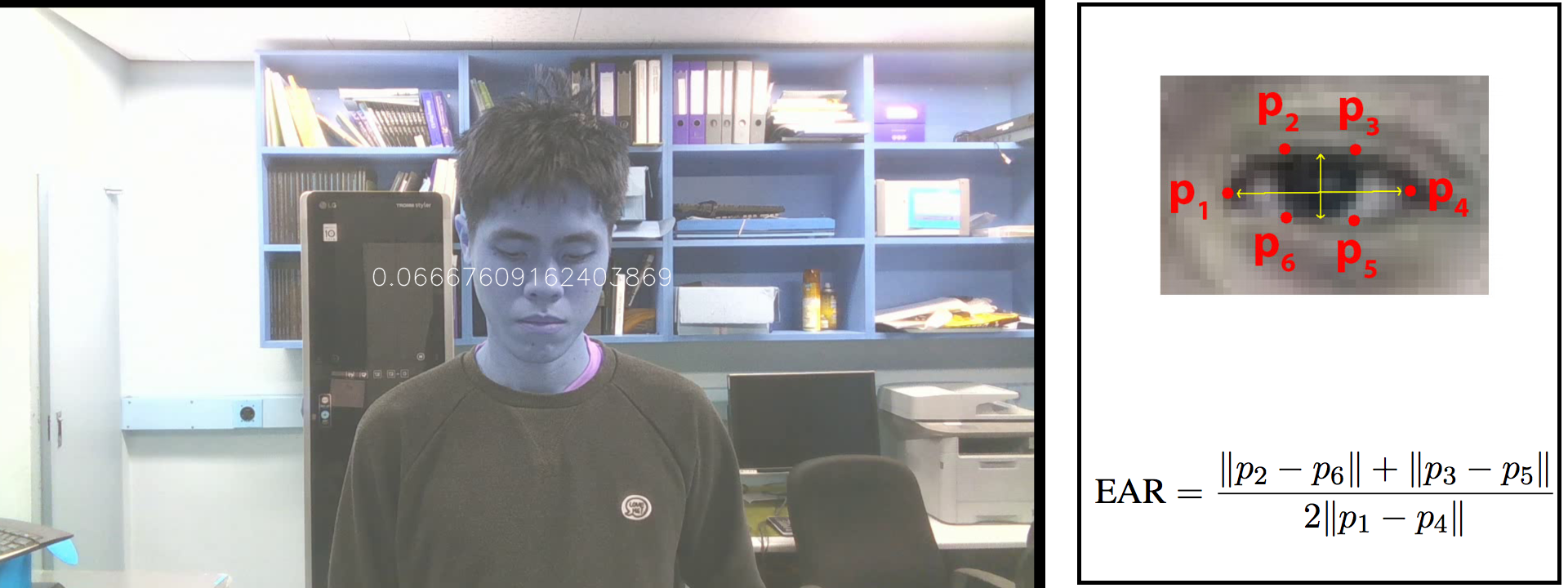}
\caption{A person looks downward to the bottom of the shelf. His eye is widely opened, but the camera position is set up inappropriately (about 24.5 cm from the top of the shelf) resulting in the camera looks at him with mostly closed eyes. We experimentally detect facial landmarks \cite{sagonas2016300,sagonas2013semi,sagonas2013300} to calculate Eye Aspect Ratio (EAR) proposed in \cite{cech2016real}. This EAR characterizes how eyes are largely or small opened.  In \cite{cech2016real}, the authors determine that eyes are widely opened with EAR above 0.2. If we inappropriately set up the camera, we can obtain EAR 0.0667 with open eyes as shown on the left side.} \label{fig3}\end{center}
\end{figure}

Fig.~\ref{fig3} shows an example of nonoptimal camera position (24.5 cm from the top of the shelf) where $\alpha_1 \neq \alpha_2$. A person who looks downward with his open eyes. However, the camera determines his eyes are mostly closed. We (humans) tend to open our eyes wider while looking upward and oppositely narrower while looking downward. This unwanted effect leads to our eye gaze algorithm (deep learning algorithm) misunderstands that his eyes are closed. The reason is that we train our deep learning model with a dataset with eyes straightly look at a camera. It has no idea to determine whether eyes look downward and are widely opened. We experimentally detect facial landmarks \cite{sagonas2016300,sagonas2013semi,sagonas2013300} to calculate Eye Aspect Ratio (EAR) proposed in \cite{cech2016real}. This EAR characterizes how eyes are largely or small opened.  In \cite{cech2016real}, the authors determine that eyes are widely opened with EAR above 0.2. If we inappropriately set up the camera, we can obtain EAR 0.0667 with open eyes as shown on the left side.

In Fig.~\ref{fig2}, $BC=138$ cm is the height of the white large paper, $750\leq d\leq 1500$ cm is the distance between the shelf and a person, $h$ is the height of a person minus 4.8 cm (approximation distance from eyes to the top of head). $b=181$ cm is the height of the shelf. We apply one property of bisection, if $\alpha_1=\alpha_2$ then $BD:DC = AB:AC$. In this equality, we have already known BC, while AB and AC can be calculated by using Pythagorean theorem. Given $h$ is the height of a person, we look to find a position $d$. The height follows Gaussian distribution with the global mean height for 159 cm for women and men 171 cm \cite{roser2013human}. From $\frac{DC}{BD} =\frac{AC}{AB}$, we have $\frac{AC}{AB} +1= \frac{BC}{DB}$. Using this fact, we put (\ref{eq1}), (\ref{eq2}) altogether and have equation (\ref{eq3}). It has three variables $d,h$ and $DB$, ($b-BC = 43$). We randomly generate $h$ with its Gaussian distribution (mean is 165, standard deviation is 6), and $d$ is uniformly distributed in the range $[750, 1500]$ and obtain $DB$ is optimal at $55.5$ cm from the top of the shelf.

\begin{equation}
\label{eq1}
    AB=\sqrt{d^2+(b-h)^2}.
\end{equation}

\begin{equation}
\label{eq2}
    AC = \sqrt{d^2 + (h-(b-138))^2}.
\end{equation}

\begin{equation}
\label{eq3}
    DB=\frac{138\sqrt{d^2 + (h-43)^2}}{\sqrt{d^2+(b-h)^2}+\sqrt{d^2 + (h-43)^2}}.
\end{equation}

If the camera optimally places at 55.5 cm from the top of the shelf, Table~\ref{tb1} depicts that a person with a specific height must stay far from the shelf to increase gaze estimation accuracy. For example, a person with 170 cm height (we ignore a space from eyes to the top of the human head for simplifying explanation) is recommended to stay away from the shelf at 1114.053 cm.

\begin{table}
\caption{Camera is placed 55.5 cm from the top of the shelf. A person with 1800 cm height is recommended to stay away from the shelf at 1212.1 cm.}
\label{tb1}
\begin{center}
\begin{tabular}{|c|c|c|}
\hline 
Item & Person Height & Distance from the shelf \\ 
\hline 
1 & 1500 & 851.453\\
\hline 
2 & 1550 & 928.174\\
\hline 
3 & 1600 & 996.515\\
\hline 
4 & 1650 & 1058.101\\
\hline 
5 & 1700 & 1114.053\\
\hline 
6 & 1750 & 1165.182\\
\hline 
7 & 1800 & 1212.100\\
\hline 
\end{tabular} 
\end{center}
\end{table}

\begin{figure}
\includegraphics[width=0.487\textwidth]{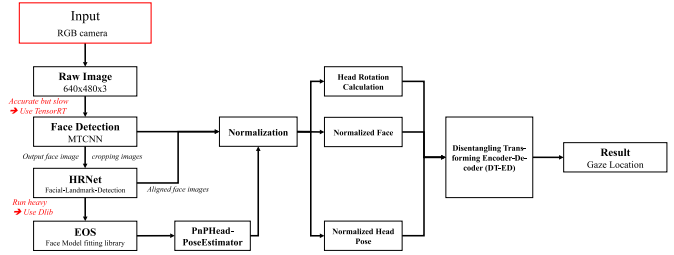}
\caption{Workflow of our gaze estimation system. It is largely inspired by FAZE proposed in \cite{park2019few}. We use MTCNN \cite{zhang2016joint} for face detection and HRNetv2 \cite{sun2019high} for facial landmark detection in fine-tuning. In inference, these two tasks are computationally expensive for edge devices. Hence, we replace the MTCNN module with a similar one but implemented in TensorRT \cite{vanholder2016efficient}, TensorRT MTCNN. Though HRNetv2 has high accuracy, we replace it with a landmark detection module in Dlib \cite{kazemi2014one,king2009dlib}.} \label{fig4}
\end{figure}

We use few shot learning techniques to reduce time to fine-tune our deep gaze estimation network. These techniques make our gaze estimation module applicable in real-world applications where we do not have much data to train. It also supports us utilize existing eye gaze research in our configured environment. We intensively inherit FAZE proposed in \cite{park2019few} to build our gaze estimation network. Fig.~\ref{fig4} shows the overall workflow of FAZE. They use MTCNN \cite{zhang2016joint} for face detection and HRNet \cite{sun2019high} for facial landmark detection. These two tasks are computationally expensive. Hence, we replace the MTCNN module with a similar one but implemented in TensorRT \cite{vanholder2016efficient}, TensorRT MTCNN Face Detector. Though HRNetv2 has high accuracy, we replace it with a landmark detection module in Dlib \cite{kazemi2014one,king2009dlib}. Facial landmark detection in Dlib is fast, which is fairly real-time without degrading accuracy compared with HRNetv2. To maintain high accuracy of inference, we still use the original HRNetv2 and MTCNN to finetune FAZE model. We only use these two replacement modules in inference on edge devices. We use data normalization proposed in \cite{zhang2018revisiting} to cancel out the significant variability in head pose such as head rotation with respect to the camera.

\begin{figure}
\includegraphics[width=0.487\textwidth]{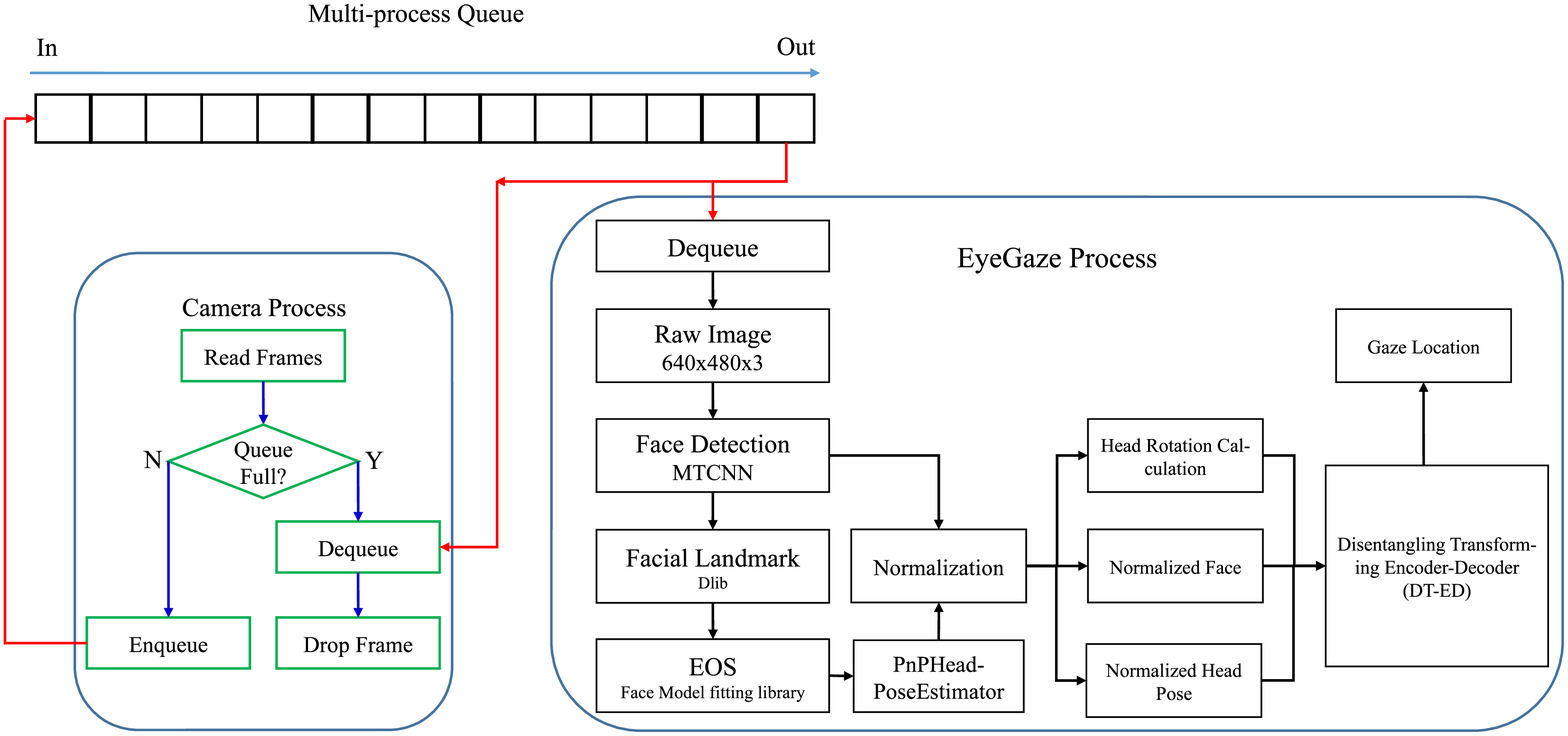}
\caption{Working with multiprocessing. Our algorithm cannot process to find gaze location in real-time (30 FPS). We use a queue with two processes. One process named “\textit{Camera Process}” puts images captured from a camera into the queue. Another process named “\textit{EyeGaze Process}” obtains the latest recent images in that queue and starts processing to estimate gaze location} \label{fig5}
\end{figure}

Despite our great effort, we cannot achieve real-time processing (30 FPS) in edge devices such as Jetson TX2. It can only run 10 to 15 FPS. Hence, we use multiprocessing and queue to ignore frames from the camera. Suppose our gaze estimation algorithm works on frame $f_i$, it moves to work on the next latest frame capture from the camera $f_j$ ($i>0, j>i+1, i,j \in \mathbb{N}$). All frames \{$f_{i+1},...,f_{j-1}$\} captured during processing frame $f_i$ are discarded. This procedure ensures that we always capture changes in a person's head such as rotation. Fig.~\ref{fig5} illustrates our idea of using a queue. We use a queue with two processes. One process named “\textit{Camera Process}” puts images captured from a camera into the queue. Another process named “\textit{EyeGaze Process}” obtains the latest recent images in that queue and starts processing to estimate gaze location. 

\section{Experiments}
\label{experiment}
\begin{figure}
\includegraphics[width=0.487\textwidth]{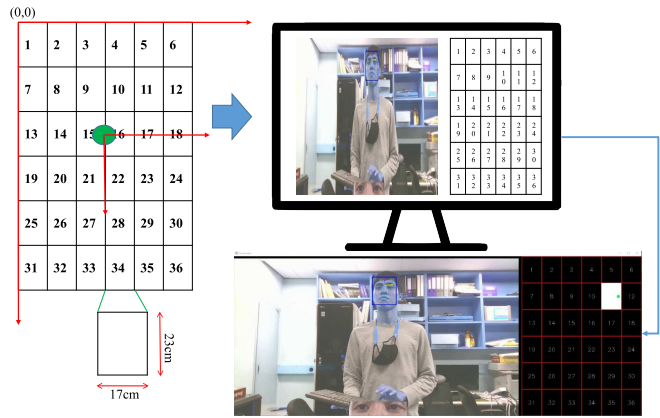}
\caption{Mapping from the physical world to screen monitor for visualization. We use the coordinates axis (0,0) at the top-left corner of the shelf to label location of each point center at each rectangle. However, our algorithm only works with the coordinates axis (0,0) at the camera pinhole location (large green dot).} \label{fig6}
\end{figure}

We put a large white paper into the shelf to simulation our real-world shopping environment as shown in Fig.~\ref{fig1} and Fig.~\ref{fig2}. It has a width of 102 cm and a height of 138 cm. We divide it into 36 equal rectangles size 17 cm in width and 23 cm in height. The camera is always at the center of width and down from the top at 55.5 cm (calculated optimal position in Section \ref{system}) as shown in Fig.~\ref{fig6}. We use the coordinates axis (0,0) at the top-left corner of the shelf to label the location of each point center at each rectangle. However, our algorithm works with the coordinates axis (0,0) at the camera pinhole location. We obtain our custom dataset to fine-tune our gaze estimation module by asking a user constantly looking at each point centered at each rectangle. In this experiment, we only have one participant. The paper (size 102x138 cm) is mapped to a screen monitor and still maintains the width height ratio. This mapping supports us to visualize our process on the monitor while collecting dataset. 

More specifically, we make a Python script to output one video file and one Pickle file \cite{pilgrim2009dive} contains ground truth (with the origin of the coordinates axis (0,0) at the top-left corner of the shelf). A person needs to look at 36 points (centered at each rectangle) to collect a calibration dataset.  We select ten frames at each point. These ten frames contain images that specify a person constantly looks at one point (out of 36). Hence, we have a total of 360 frames in one calibration video. However, to fine-tune our deep learning model, we only use three random images for training and one image for validation from those ten collected images of a given point. In Fig.~\ref{fig6}, there have thirty-six labeled squares ranging from 1 to 36. We experimentally find that fine-tuning and validation locations should be equally distributed in the shelf in order to avoid bias in any region (left, right, top, down) respect to camera location. Hence, We use squares with number \{8, 11, 26, 29\} for validation. Those points location are equally located with respected to camera location. The rest 32 points are used in different sets of fine-tuning as shown in Table~\ref{tb2}. 

Our implementation is intensively based on a few shot learning \cite{park2019few}. We ignore the training phrase since their model was fairly trained with the following number of calibration points \{1, 2, 3, 4, 5, 6, 7, 8, 9, 10, 11, 12, 13, 14, 15, 16, 17, 18, 32, 64, 128, 256\}. We do not re-train their network and only use their pre-trained weight on inference. Logically, gaze estimation accuracy can be improved if we increase the number of calibration points. However, we do not have enough time and labor resources to label our custom dataset. Hence, we use different sets of points to find the reasonable number of calibration points shown in Table~\ref{tb2}. As mentioned earlier, we fix four points to validate the fine-tuning process. 

\begin{figure}
\includegraphics[width=0.4893\textwidth]{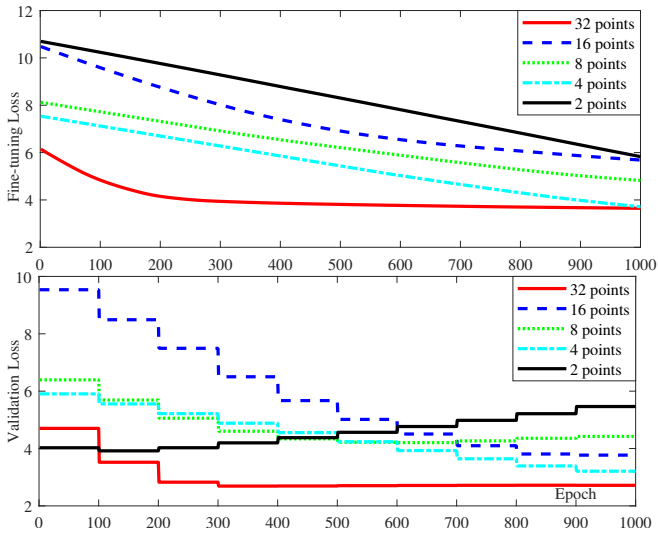}
\caption{We fine-tune FAZE network using five different sets of points (located at each rectangle) shown in Table~\ref{tb2}. If we fine-tune with two points, validation loss increases compared to other sets. Hence, each person needs at least four samples to fine-tune FAZE network. As we increase the number of samples (points) to fine-tune the gaze estimation network, validation loss reduces accordingly.} \label{fig7}
\end{figure}

Fig.~\ref{fig7} illustrates fine-tuning loss (training and validation) with respect to an epoch. If we fine-tune with two points, validation loss increases compared to other sets. Hence, each person needs at least four samples to fine-tune FAZE network. As we increase the number of samples (points), to fine-tune the gaze estimation network, validation loss reduces accordingly.

\begin{table}
\caption{Difference training (fine-tuning) sets of points. Each point centered at each rectangle.} 
\label{tb2}
\begin{center}
\begin{tabular}{|P{0.3cm}|P{1.6cm}|p{5.5cm}|}
\hline 
Item & Tuning points & Point indexes \\ 
\hline 
1 & 2 & 6, 31\\
\hline 
2 & 4 & 3, 13, 18, 33\\
\hline 
3 & 8 & 1, 3, 6, 13, 18, 31, 33, 36\\
\hline 
4 & 16 & 1, 3, 4, 6, 13, 15, 16, 18, 19, 21, 22, 24, 31, 33, 34, 36\\
\hline 
   &    & 1,  2,  3,  4,  5,  6,  7,  9, 10, 12, 13, 14, 15, 16, 17, \\
5  & 32 & 18, 19, 20, 21, 22, 23, 24, 25, 27, 28, 30, 31,\\
 & & 32, 33, 34, 35, 36\\
\hline 
\end{tabular} 
\end{center}
\end{table}

\begin{figure}
\includegraphics[width=0.487\textwidth]{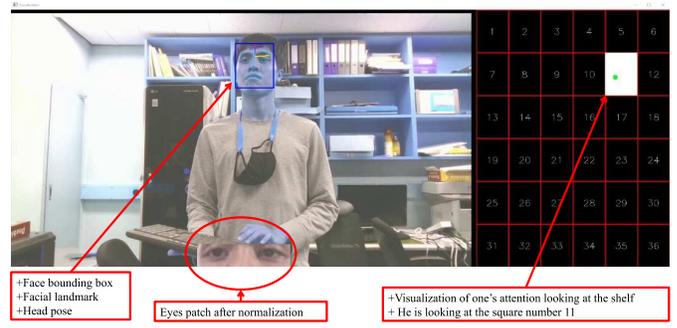}
\caption{We display the appearance of a person with (1) face bounding box \cite{zhang2016joint}, (2) facial landmark \cite{sun2019high}, (3) head pose, (4) eyes patch after normalization \cite{zhang2018revisiting}. We also output (5) the location of where the user gazes at. For example, a user attentively looks at the square number eleven.} \label{fig8}
\end{figure}

Fig.~\ref{fig8} visually shows the output while fine-tuning our gaze network. RGB images are feed into our system. Our system outputs (1) face bounding box \cite{zhang2016joint}, (2) facial landmark \cite{sun2019high}, (3) head pose, (4) eyes patch after normalization \cite{zhang2018revisiting}. We also output (5) the location of where the user gazes at. For example, a user attentively looks at the square number eleven.

\begin{figure}
\includegraphics[width=0.4893\textwidth]{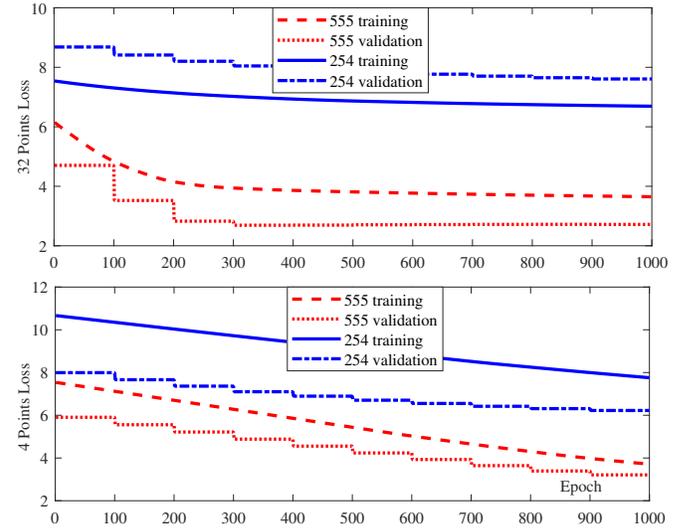}
\caption{We set up an RGB camera to capture frames from two different locations centered along the width. (1) optimal position 55.5 cm, and non-optimal position 24.5 cm from the top of the shelf. Fine-tuning results with two sets of 32 and 4 points from Table~\ref{tb2}. It shows that the camera at position 25.4 cm cannot reduce training and validation loss even though we increase the number of samples from 4 to 32.} \label{fig9}
\end{figure}

\begin{figure}
\includegraphics[width=0.487\textwidth]{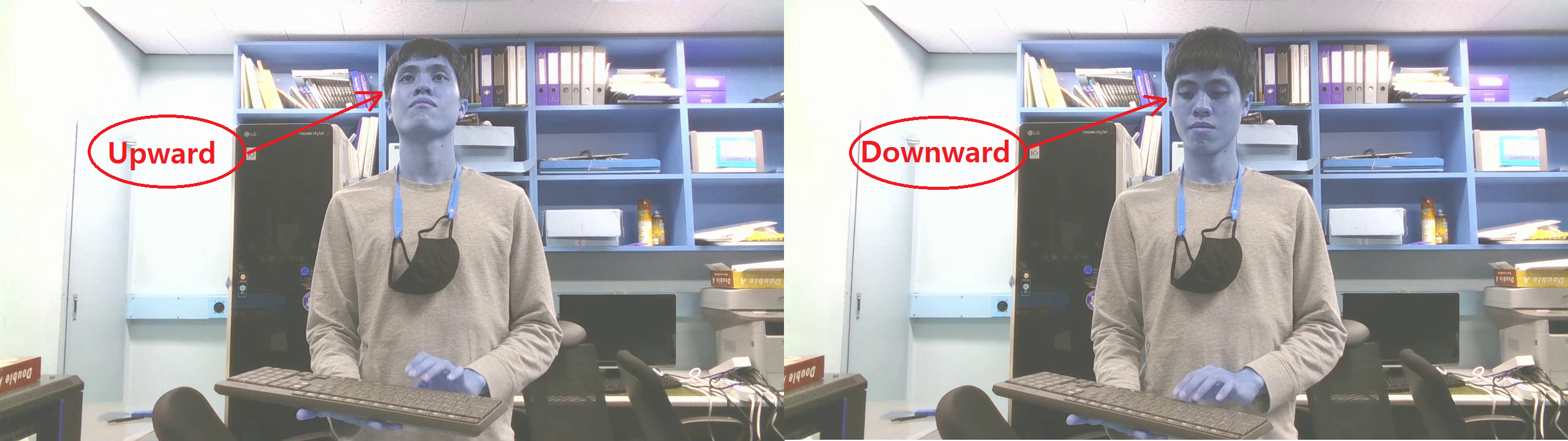}
\caption{We (humans) tend to open our eyes wider while looking upward and oppositely narrower while looking downward. We have found the optimal position of the camera (55.5 cm from the top) but this effect prevents us to equally capture opened eyes when looking upward and downward. However, we have overcome the effect and estimate eyes gaze more accurately compared to what we can obtain in Fig.~\ref{fig3}.} \label{fig10}
\end{figure}

If we set up a camera to capture frames inappropriately, we cannot archive high-accuracy gaze estimation. As shown in Fig.~\ref{fig9}, we set up an RGB camera to capture frames from two different locations. We compare training and validation loss of two camera positions which are 55.5 and 25.4 cm from the top of the shelf. It shows that the camera at position 25.4 cm cannot reduce training and validation loss even though we increase the number of samples from 4 to 32. Especially, fine-tuning with 32 points even increases validation loss and maintains at 8. Oppositely, the camera at position 55.5 cm logically reduces training and validation loss to 2. Despite our hard efforts in finding a camera position, we still face the eyes-looking effects shown in Fig.~\ref{fig10}. Though, we mathematically find the optimal camera position 55.5 cm from the top. The eye-looking effect still prevents us to obtain equally opened eyes when looking upward and downward but it is much better compared to what we can obtain in Fig.~\ref{fig3}. 

\begin{figure}
\includegraphics[width=0.487\textwidth]{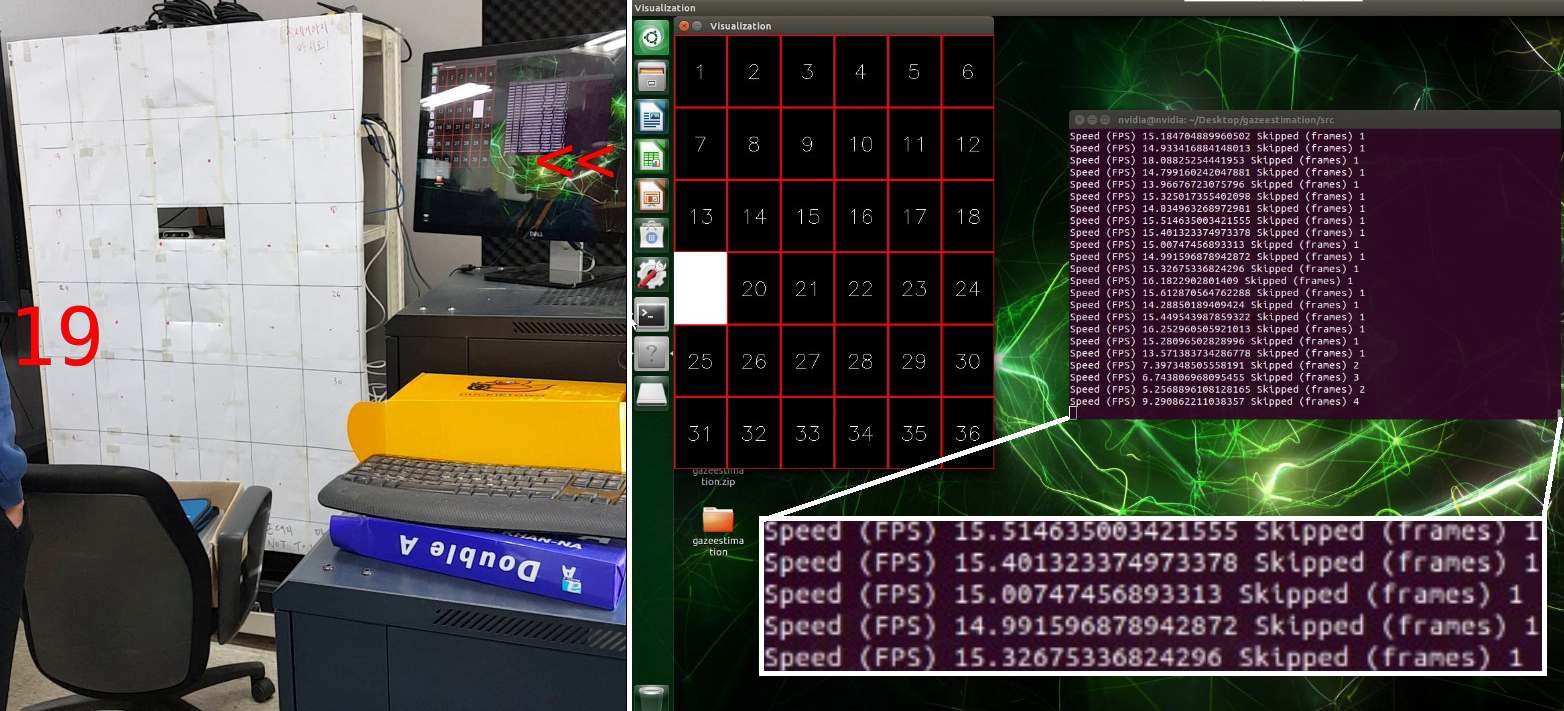}
\caption{Our experiment with the optimal camera position at 55.5 cm from the top of the shelf. A user gazes at rectangle number 19 (leftmost side). Our gaze estimation module correctly identifies the user's gazing (shown in a white rectangle). It skips one to five frames in order to process real-time images captured from the camera. The processing speed is about 10 to 15 FPS.} \label{fig11}
\end{figure}

Fig.~\ref{fig11} shows the final output of our experiment with the optimal camera position at 55.5 cm from the top of the shelf. A user gazes at rectangle number 19 (left side). Our gaze estimation module correctly identifies the user's gazing (shown in a white rectangle, right side). It skips one to five frames to process the latest images captured from the camera. A few captured images are discarded to ensure that our gaze estimation algorithm is not stuck or keeps processing past frames. Every change of personal appearance such as head rotation, eyes blink is constantly processed. The processing speed is about 10 to 15 FPS.

\section{Conclusion}
In this paper, we bring research in gaze estimation to a real-world application. Gaze estimation has been done intensively in the literature, but it has many limitations and far beyond real-world constraints. We bridge the gap to find the optimal position for the camera. This effort maintains high accuracy grasped from our reference research. Then, we implement to run state-of-the-art gaze estimation on edge devices.  Our experiment proves that setting the camera at an appropriate position supports us to obtain eyes gaze correctly. It results in maintaining high accuracy compared to the results that have been done in the research environment. However, we only consider a scenario in which only one person statically standing in front of the camera and has no movement such as walking. Furthermore, the 3D shape of the face around the eyes strongly affects the gaze estimation accuracy so that the optimal camera position may be different among different people. In the future, we intend to experimentally and quantitatively confirm our optimal camera position with data from multiple participants. Additionally, we plan multiple cameras to avoid constraints from our current research environment making it more realistic. We can also extend it to a scenario where many people move around while looking at the camera.

\label{conclusion}


\bibliography{ref}
\bibliographystyle{IEEEtran}

\end{document}